\documentclass[12pt,a4paper]{article}
\usepackage{times}
\usepackage{helvet}
\usepackage{courier}
\usepackage{url}
\usepackage{graphicx}
\usepackage{hyperref}
\usepackage{todonotes,bbm,bm,amsthm,amsmath,amsfonts,amssymb,mathtools,amssymb,mathrsfs,xfrac,tikz,pgf}

\newtheorem{example}{Example}
\newtheorem{theorem}{Theorem}

\newtheorem{corollary}{Corollary}
\newtheorem{definition}{Definition}
\newtheorem{lemma}{Lemma}
\newtheorem{proofoft}{Proof of Th.}
\newtheorem{proofofl}{Proof of Lemma}
\usepackage{graphicx}
\title{IMAGINARY KINEMATICS}
\author{Sabina Marchetti\\ Sapienza University of Rome\\ Rome (Italy)\\ {\tt sabina.marchetti@uniroma1.it}
\and Alessandro Antonucci\\ IDSIA\\ Lugano (Switzerland)\\{\tt alessandro@idsia.ch}} %
\date{\today}

\begin{document}
\maketitle
\begin{abstract}
We introduce a novel class of adjustment rules for a collection of beliefs. This is an extension of Lewis' imaging to absorb probabilistic evidence in generalized settings. Unlike standard tools for belief revision, our proposal may be used when information is inconsistent with an agent's belief base. We show that the functionals we introduce are based on the \emph{imaginary} counterpart of \emph{probability kinematics} for standard belief revision, and prove that, under certain conditions, all standard postulates for belief revision are satisfied.
\end{abstract}
\section{Introduction}
The theory of belief revision, originated in the work of Alchourr\'{o}n, G\"{a}rdenfors and Makinson \cite{AGM}, is aimed to maintain consistency of a knowledge base when updated information is gathered to a rational agent, or \emph{You}. In the present work we will focus on the probabilistic framework, where Your knowledge base is represented by a (closed and convex) collection of probability mass functions, and some observational process is expected to induce an adjustment in the model.\footnote{Here we intend an \emph{adjustment} as a generalized \emph{updating}. We avoid this latter term as in the literature it is often intended as equivalent to conditioning.} With probabilities, evidence on some variables is called \emph{inconsistent} when it contradicts certainty (or impossibility) in Your knowledge base. We provide an example to motivate our contribution.
\begin{example}\label{celeste}
While swimming in a lake, Celeste sees some black birds from the distance. She knows black birds living around that lake are rather tame, while swans might be very aggressive. She is also sure that only white or grey swans exist, although the birds she sees actually look like swans. While reasoning about that, a sailor informs her that a small group of black swans has been spotted around the area. Should Celeste be worried about the birds she sees?
\end{example}
Classic belief revision operators, introduced in Section~\ref{back}, fail to absorb information from an observational process when inconsistencies arise such as in Example~\ref{celeste}. This feature was motivated in the literature by a \emph{partiality} principle \cite{cozic2011imaging}, discussed below. Still, a rule for the adjustment of a model to any piece of evidence ought to be required by a rational agent, to avoid building a new model from scratch when unexpected information shows up. Such an operator ought to update the knowledge base to be consistent with new evidence, while leaving previous beliefs on related events as unchanged as possible. We will characterize optimality requirements for such adjustment operators as an \emph{imaginary kinematics} in Section~\ref{ima}, and extend them to deal with generalized forms of evidence. Particularly, we consider probabilistic evidence, and extend it to i) conditional assessments, and ii) \emph{imprecise} assessments, that may be intended as originating from a qualitative judgment. Section~\ref{ope} will introduce adjustment functionals based on Lewis' imaging, and study their features and properties. We will refer throughout to partial operators as \emph{revision} rules, as opposed to general \emph{adjustment} ones.
\section{Background}\label{back}
Let $\Omega$ be any space of atoms - atomic (Boolean) propositional variables - and let a \emph{world} $\omega$ be any assignment of truth to each element from $\Omega$, such that there exist up to $2^{|\Omega|}$ \emph{conceivable} worlds.\\
Any propositional formula $\phi\in\mathcal{L}$, countable set of all formulae on $\Omega$, is satisfied by worlds in $[\phi]\subseteq\Omega$. Formally, when $\omega$ satisfies $\phi$ we write $\omega\models\phi$; that is, $\omega\in[\phi]$ if and only if $\omega\models\phi$. Logical connectives $\{\land,\lor,\neg\}$ - conjunction, disjunction and negation, respectively - may be used to concatenate several formulae. Also, $\top$ and $\perp$ denote, respectively, tautology and contradiction.\\
A rational agent (or You) is equipped with a collection of belief states over some $A\subseteq\Omega$, whose elements may be equivalently defined by closed sets of formulas in a propositional logic language. Formally, a belief state over the set of all \emph{conceivable} worlds $A\subseteq\Omega$, is represented by a \emph{probability mass function} (PMF) $P_A$, defined as follows:
\begin{align*}
P_A(A)= \left\{(\omega,P(\omega)): \begin{array}{ll}
P(\omega)\geq 0,\omega\in A, \\
\sum_{\omega\in A} P(\omega)=1
\end{array}\right\} \,. 
\end{align*}
Granular belief $P_\Omega$ is similarly defined with respect to every $\omega\in\Omega$. We just write $P$, when the domain is clear from the context.\\
Let $\mathbf{X}$ be a collection of $n$ discrete variables, $n\geq 1$, $\omega$ corresponds to $\mathbf{x}$, configuration of $\mathbf{X}$ in its joint possibility space, and $\Omega \equiv \Omega_\mathbf{X}$, while $\mathcal{L}$ reduces to a collection of statements $\{\phi \bowtie c: \phi\in\mathcal{L},\bowtie \in\{=,\geq,\leq\},c\in[0,1]\}$. Also, $A$ represents any arbitrary tautology, such that any $P_A$ is strictly positive on $A$ (and contains zero elements only otherwise). For a given formula $\phi$,
\begin{equation*}
P(\phi) = \sum_{\mathbf{x}\in\Omega:\mathbf{x}\sim A}P(\mathbf{X}=\mathbf{x})\mathbb{I}_{\mathbf{x}\models\phi}\,,
\end{equation*}
with $\sim$ denoting consistency among events. E.g., let $n=3$, $\phi=\{x \land \neg y\}$, $(x,\neg y,z)\sim [\phi]$, whatever $z$ in $\Omega_Z$, coarse partition of $\Omega$ induced by variable $Z$. For the sake of brevity, in the following, we write $P(\mathbf{x})$, rather than $P(\mathbf{X}=\mathbf{x})$.\\
In the general case, a collection of deductively closed set of propositions, i.e., belief states, may be used to specify a \emph{credal set} (CS) $K(\mathbf{X})$. Any CS $K$ is defined by a set of linear constraints, and may be equivalently characterized as the convexification of its extreme points, denoted as $\mathrm{ext}[K]$. Let $K_1$ and $K_2$ be any two CSs over $\mathbf{X}$, they are \emph{equivalent}, $K_1\equiv K_2$, if and only if $\mathrm{ext}[K_1]=\mathrm{ext}[K_2]$. For each $x\in\Omega_X$, $\underline{P}(x)=\min_{P(x)\in \mathrm{ext}[K(X)]}P(x)$ (and $\overline{P}(x)=\max_{P(x)\in \mathrm{ext}[K(X)]}P(x)$) corresponds to the lower (and upper) envelope of CS $K(X)$, for any $X\in\mathbf{X}$. See \cite{walley1991statistical} for details on CSs. We refer to \emph{sharp} or \emph{imprecise} probabilities to distinguish between $|\mathrm{ext}[K]|=1$ and $|\mathrm{ext}[K]|>1$, respectively.\\
$K^\Phi$ denotes the subset of belief states in $K$ that satisfy a collection of formulae $\Phi$. Any belief state satisfies $\Phi$, i.e., $P\models \Phi$, whenever it holds $P\models\phi$, for each $\phi\in\Phi$. Any set $\Phi$ is \emph{accepted} whenever it is consistent with each $P\in K$, it is \emph{rejected} if its negation only, $\neg\Phi$, is, or it is \emph{neutral} if both are consistent. Let $c\in[0,1]$, for a given formula $\phi$, $P\models \left(\phi\bowtie x\right)$ whenever $P(\phi) \left(=\sum_{\mathbf{x}\sim[\phi]} P_A(\mathbf{x}) \right) \bowtie c$, $\bowtie\in\{=,\leq,\geq\}$.\\
For a given belief set, three main operations are relevant to adjust it to satisfy any given $\phi$. These are contraction, expansion and revision from AGM theory \cite{AGM}, whose consistency postulates are mostly known from the KM reformulation in \cite{katzuno1992propositional}. Suppose an agent's knowledge base is represented by a CS $K$ over $\mathbf{X}$, and let $\phi$ be any upcoming formula, such that adjustment of $K$ by $\phi$ is operated by $\circ$. Katzuno and Mendelzon's postulates translate as follows:
\begin{description}
\item[KM1] $(K \circ \phi) \models \phi$,
\item[KM2] Let $K\models \phi$, $(K\circ  \phi) \equiv (K\cup  \phi)$,
\item[KM3] If $\phi\neq\perp$, then $\left(K\circ \phi\right)\neq \perp$,
\item[KM4] If $K_1\equiv K_2$ and $\phi_1\equiv \phi_2$, then $\left( K_1 \circ \phi_1\right) \equiv \left(K_2 \circ \phi_2\right)$,
\item[KM5] If $\left(K \circ \phi\right) \models \psi$, then $\left(K \circ \left(\phi \land \psi\right)\right)$, for any further formula $\psi$,
\item[KM6] If $(K \circ \phi) \models \psi$, then $\left(K \circ \left(\phi \land \psi\right)\right)$ implies $\left(\left(K \circ \phi\right) \models \psi\right)$.
\end{description}
Any operator $\circ$ that satisfies all KM postulates is equivalent to a revision process based on total pre-orders \cite{katzuno1992propositional}.

AGM postulates, and their KM formulation, have been followed by a massive literature on their limitations and possible extensions. Two major shortcomings of AGM theory arise when revision involves conditional formulae \cite{douven2011}, and in the iterated setting \cite{goldszmidt1992qualitative}. See also \cite{darwiche1997} on additional postulates for iterated belief revision.

In the classical probabilistic framework, $K(\mathbf{X})$ is made by a single PMF, that is $\mathrm{ext}[K(\mathbf{X})]=\{P(\mathbf{X})\}$. When one or more elements from $\mathbf{X}$ are observed, $P$ is adjusted, i.e., updated, accordingly by standard \emph{conditioning}. Let $\alpha$ be any event from $\Sigma$, the $\sigma$-algebra induced by $\Omega$, and suppose $(X=x)$ with $x\in\Omega_X$ and $X\in\mathbf{X}$, is observed and such that $P(x)>0$, it holds:
\begin{equation}\label{conditioning}
P(\alpha|x)=P(\alpha, x)/P(x) \,.
\end{equation}

A (marginal) probabilistic observation corresponds to a PMF over the countable possibility space of variable $X\in\mathbf{X}$. Such evidence bears an impression of the degree of reliability that is associated to each (forecasted) event, i.e., on the \emph{evidence of uncertainty} \cite{peng2010bayesian}. We define probabilistic evidence as some PMF $P'_X$ over $\Omega_X$, such that $P(x) \not= P_X'(x)$ for some $x \in \Omega_X$. It corresponds to the collection of formulae $\Phi_X$, whose generic element is $\phi_x=(\{x\}=c_x)$, $c_x\in[0,1]$, $x\in\Omega_X$, with $\sum_{x\in\Omega_X}c_x=1$. $P'_X$ may be intended as a set of probabilistic constraints on the system modeled by $P$ \cite{decampos}. A general adjustment operator is the functional $\circ$, mapping any $P$ to $P^\circ$, such that $P^\circ\models P_X'$. By the partiality principle mentioned above, standard revision of $P$ by $P_X'$ requires preservation of zero-probability events. Rationality of partiality has been advocated by several authors (e.g., \cite{dietrich2016}). The intuition is the following: Your beliefs ought to be calibrated with available evidence, if any. This way, certainty on the occurrence of event $(X=x')$ requires $P(x)=0$, for each $x\neq x'$ in $\Omega_X$. If You accepted to change Your mind on $(X=x)$, then You would rather be reasonably sure about its non-occurrence, rather than certain; but then $P(x)\neq0$. As a consequence, certainty on the occurrence of an event, say $x$, implies certainty to $P'_X$, since $P'_X(x')$ is floored to zero by every $x' \not= x$ in $\Omega_X$.

Kinematical mechanics for the adjustment of a belief set are intended as consistency principles, that we are willing to choose over a purely \emph{minimal distance} based approach \cite{boutilier1996iterated}. We introduce \emph{probability kinematics} following Wagner's characterization \cite{wagner2002probability}.

\begin{definition}[Probability kinematics \cite{jeffrey1965, wagner2002probability}]\label{def:pk}
Let $P$ and $P^\circ$ be any two PMFs over $(\Omega,\Sigma)$, and let $\Omega_X$ be a countable collection of pairwise disjoint events in $\Sigma$, i.e., a coarse partition of $\Omega(\equiv\Omega_\mathbf{X})$. $P^\circ$ comes from $P$ on $\Omega_X$ based on probability kinematics (PK) if there exists a sequence $P_X'(X)=\{{P}_X'(x):x \in \Omega_X, \sum_{x\in\Omega_X} P_X'(x)=1\}$ such that it holds:
\begin{description}
\item[PK1] $P^\circ(\alpha | x) = P(\alpha |x)$, for each $x\in\Omega_X$,
\item[PK2] $P^\circ(X)=P_X'(X)$,
\end{description}
for any event $\alpha\in\Sigma$.
\end{definition}

In words, $P$ is changed to agree with $P_X'$ (PK2), while preserving relevance of each $x\in\Omega_X$ to any event $\alpha\in\Sigma$ (PK1).\\
An equivalent characterization of PK yields the well-known Jeffrey's rule:
\begin{definition}[Jeffrey's Rule \cite{jeffrey1965}]
Let $P$, $P^\circ$ and $P_X'$ as above. Jeffrey's rule ($\circ_J$) adjusts $P$ to satisfy ${P}_X'$:
\begin{equation*}
\left(P \circ_J {P'}_X\right)(\alpha) = \sum_{x \in \Omega_X} P(\alpha, x) \frac{{P'}_X(x)}{P(x)}
\end{equation*}
We denote the Jeffrey's revision of $P$ on $\Omega_X$ as $P_X^{\circ_J}$.
\end{definition}

Deterministic knowledge on event $(X=x)$ may be specified by $P'_X(X)$ such that $P_X'(x) = 1$ at $x$ and zero otherwise.\footnote{While probabilistic findings extend standard evidence, they do not necessarily result from an observation process. E.g., they may be gathered as forecasts produced by external sourced whose system of knowledge is not disclosed (e.g., betting odds), or qualitative evaluations from experts. Thorough characterization of uncertain evidence has been provided in the survey of \cite{mrad2015explication}, and related works. There, probabilistic evidence is further distinguished into fixed and not-fixed. Such distinction is critical to iterated belief revision.} It holds:
\begin{equation}\label{eq:jandc}
\left(P \circ_J P_X'\right)(\alpha) \equiv P(\alpha|x) \,,
\end{equation}
where the righ hand-side is just conditioning from Eq.~\eqref{conditioning}. Such \emph{hard} evidence \cite{valtorta2002} trivially corresponds to $\phi=\{x\}$, $x\in\Omega_X$.

Suppose evidence is gathered conditional on some variable $Y$ taking value $y\in\Omega_Y$. We define conditional (probabilistic) evidence as the collection of probabilistic statements $P_{X|y}'(X|y)$, such that $P_{X|y}'(x|y)\geq0$, for each $x\in\Omega_X$, and $\sum_{x\in\Omega_X}P_{X|y}'(x|y)=1$, provided $P(y)>0$. Equivalently, $\Phi_{X|y}$, with generic element $\phi_{x|y}=(\{y \to x\}=c_x)$, with $\sum_{x\in\Omega_X}c_x=1$. A kinematical revision rule would require the following conditions to hold:

\begin{definition}[Conditional PK \cite{bradley2005}]\label{def:copk} Let $P$ and $P^{\circ}$ be any two PMFs on $(\Omega,\Sigma)$. Let $P(y)>0$, $P^\circ$ comes from $P$ on $\Omega_X\times\{Y=y\}$ based on conditional probability kinematics (CPK) if there exists a sequence $P_{X|y}'(X|y)$ as above such that it holds:
\begin{description}
\item[CPK1] $P^\circ(\alpha|x,y)=P(\alpha|x,y)$, for each $x\in\Omega_X$,
\item[CPK2] $P^\circ(\alpha|y')=P(\alpha|y')$, for each $y'\in\Omega_{Y}\backslash\{y\}$,
\item[CPK3] $P^\circ(Y)=P(Y)$,
\item[CPK4] $P^\circ(X|y)=P_{X|y}'(X|y)$.
\end{description}
\end{definition}

The following operator may be used to revise $P$, extending Jeffrey's rule to the conditional setting:
\begin{definition}[Adams' Conditioning \cite{bradley2005,douven2011}]
Let $P$, $P^\circ$ and $P_{X|y}'$ as above, with $P(y)>0$. Operator $\circ_A$ yields the Adams' revision ($P_{X|y}^{\circ_A}$) of $P$ that is consistent with ${P}_{X|y}'$ if it is obtained as:
\begin{align*}
&\left(P \circ_A {P}_{X|y}'\right)(\alpha) =\\
&P(\alpha,\neg y) + \sum_{x \in \Omega_X} P(\alpha, x, y) \frac{{P}_{X|y}'(x|y)}{P(x|y)} \,.
\end{align*}
\end{definition}
By \cite[Th.5]{bradley2005}, Adams' conditioning yields the unique PMF that satisfies CPK1-CPK4.
Let us consider that in the running example.
\begin{example}[Ex.~\ref{celeste} continued]
Celeste's beliefs are formalized as follows: let $\Omega_Y=\{y\equiv\text{Swan},\neg y\equiv\neg\text{Swan}\}$, $\Omega_X=\{x_W\equiv\text{White},x_G\equiv\text{Grey},x_B\equiv\text{Black}\}$ and $\Omega_Z=\{z\equiv\text{Aggressive},\neg z\equiv\text{Tame}\}$.\\
It holds:
\begin{align*}
P(Y)=\left\{(y, 0.7),(\neg y, 0.3)\right\} \,,
\end{align*}
\begin{align*}
P(X|Y)=\left\{\begin{array}{ll}
(x_W|y, 0.8), (x_G|y, 0.2),\\
(x_B|y, 0),(x_W|\neg y, 0.5),\\
(x_G|\neg y,0.3),(x_B|\neg y, 0.2)
\end{array}\right\} \,,
\end{align*}
\begin{align*}
P(Z|Y)=\left\{\begin{array}{ll}
(z|y, 0.95), (\neg z|y, 0.05),\\
(z|\neg y, 0.2),(\neg z|\neg y, 0.8)
\end{array}\right\}\,.
\end{align*}
According to Celeste's beliefs, $P(z|x_B)=0.2$. Based on the sailor's words, Celeste is willing to adjust her beliefs to be consistent with $P_{X|y}'(X|y)=\{(x_W, 0.8), (x_G, 0.1),(x_B, 0.1)\}$. Straightforward application of Adams' conditioning is undefined, since $P(x_B|y)=0$, while $P_{X|y}'(x_B|y)\neq0$. The same would occur with simple Jeffrey's rule, if any $P_X'(x)\neq0$ was provided, given $P(x)=0$, for some $x\in\Omega_X$. How could Celeste incorporate such reliable knowledge in her beliefs?
\end{example}

\emph{Imaging} was introduced by \cite{lewis1976probabilities} as a non-trivial alternative to conditioning on inconsistent events. Roughly, it represents the ``\emph{thought experiment by a minimal action}'' \cite{fusaoka2003linear} that makes a formula consistent.\\
Going back to the propositional language, if some world $\omega$ is inconsistent with formula $\phi$, according to a knowledge base, imaging shifts beliefs towards those that are closest to $\phi$, called $\phi$-worlds. $\gamma(\omega,\phi)$ is called a \emph{closest world function}, mapping $\omega$ to its closest $\phi$-world; 
see \cite{lewis1986plurality} for a detailed discussion. In our formalism, $(\phi=\{x\})$ requires $\gamma(\mathbf{x},\phi)=(\mathbf{x}\backslash\{X\},x)\in\Omega$, for any $\mathbf{x}\in\Omega$.

\begin{definition}[Imaging \cite{lewis1976probabilities}]\label{def:Imaging}
Let $P$ be any PMF over $(\Omega, \Sigma)$. For a given $\phi$ and closest world function $\gamma(\cdot,\phi)$. $P_\phi^{\circ_I}$ is the image of $P$ on $\phi$ if it is obtained by $\circ_I$ as:
\begin{align*}
\left(P \circ_I \{\phi\}\right)(\alpha)
&= \sum_{\omega' \in\alpha} \sum_{\omega\in\Omega}P(\omega) \mathbb{I}_{\gamma(\omega,\phi)=\omega'} \,.
\end{align*}
\end{definition}
In Lewis' words, by imaging on event $\phi$, \emph{``probability is moved around, but not created or destroyed''}, while \emph{``every share stays as close to it as it can to the world it was originally created''} \cite[p. 310-311]{lewis1976probabilities}. To summarize: i) inconsistent evidence is accounted for in the image of $P$, whereas conditioning is left undefined; ii) imaging changes the whole belief set to comply with reliable knowledge $\phi$, while conditioning redefines the domain of $P$, focusing on worlds in $\Omega$ consistent with $\phi$.
\begin{example}
Let $\mathbf{X}=\{X,Y\}$, with $P(x,y)=P(x,\neg y)=0$, $P(\neg x,y)=0.6$ and $P(\neg x,\neg y)=0.4$. Given $(\phi=\{x\})$, imaging on it yields $\left(P \circ_I \{X=x\}\right)(y)=0.6$, which corresponds to $P(y)$. If conditioning was applied, $P(Y|x)$ would not be defined.\\
Consider $\alpha=\{x\}$, $\left(P \circ_I \{X=x\}\right)(x)= 1$: $\circ_I$ adjusts $P$ to always be consistent with $\phi=\{x\}$.
\end{example}
Generalized forms of imaging were introduced in the literature, see, e.g., \cite{gardenfors1988knowledge,rens2016revision}. See also \cite{zhuang2017unifying} on a unifying approach to belief adjustment.

G{\"u}nther \cite{gunther2017learning} introduced Jeffrey's imaging, that we denote as $\circ_{jI}$, for the generalized case of probabilistic formula $\left(\phi= c\right)$, with $c\in[0,1]$.\footnote{G{\"u}nther's definition assumes $c\in(0,1)$.} Adjustment operator $\circ_{jI}$ trivially extends \emph{partial} imaging \cite{ramachandran2010belief}.
\begin{definition}[Jeffrey's Imaging \cite{gunther2017learning,ramachandran2010belief}]\label{def:ji}
Let $P$ be any PMF over $(\Omega, \Sigma)$. For a given formula $\left\{\phi=c\right\}$, with $c\in[0,1]$, $P_X^{\circ_{jI}}$ comes from $P$ by Jeffrey's imaging $\circ_{jI}$ on $\left\{\phi=c\right\}$ if it holds:
\begin{align*}
\left(P \circ_{jI} \left\{\phi=c\right\}\right)(\alpha)=P_\phi^{\circ_I}(\alpha)c+ P_{\neg\phi}^{\circ_I}(\alpha)(1-c)
\end{align*}
We denote the Jeffrey's image of $P$ on $\{\phi=c\}$ as $P_\phi^{\circ_{jI}}$.
\end{definition}

Both standard and Jeffrey's imaging are \emph{homomorphic} change functions (see \cite{gardenfors1988knowledge} and \cite[Obs.1]{ramachandran2010belief}, respectively), i.e., they define a structure-preserving map. A generalized characterization of Jeffrey's imaging will be provided below, within the multi-valued imprecise-probabilistic framework (see Definition~\ref{pji}).

Just like Your beliefs may be encoded by a CS $K$ on $\Omega$, probabilistic evidence may come as a (closed and convex) collection of PMFs $K_X'$ on $\Omega_X$, i.e., a CS that we call \emph{credal} (or imprecise) evidence. This latter generalizes sharp probabilistic evidence to the case $|\mathrm{ext}[K_X'(X)]|\geq 1$:
\begin{equation*}
K_X'(X)=\{P(x): \underline{P}(x)\leq P(x)\leq \overline{P}(x), x \in\Omega_X\} \,.
\end{equation*}

$K_X'$ may be equivalently specified by the collection of formulae $\phi_x=(\{x\}\bowtie c_x)$, $c_x\in[0,1]$, for each $x\in\Omega_X$, provided $\sum_{x\in\Omega_X}c_x\bowtie 1$, $\bowtie\in\{=,\leq,\geq\}$. \footnote{To guarantee $P_X'(x)\in[0,1]$, we also require $P_X'(x)\leq 0$ and $P_X'(x)\geq 1$ always reduce to equalities.}

Our contributions will tackle probabilistic belief adjustment by (possibly inconsistent) sharp or imprecise probabilities, following an approach based on the imaginary counterparts of PK. This is analogous to what has been done in \cite{ma2011,zhou2014} within the framework of evidence theory. 

Following \cite{zhou2014}, we are willing to check a further consistency requirement, that would reproduce Eq.~\eqref{eq:jandc}. In this way, any adjustment \emph{kinematical} operator reduces to some form of conditioning when probabilistic evidence strengthens to full observation.

\section{Imaginary Kinematics}\label{ima}

We lay bare the kinematical conditions that ought to be satisfied by any belief adjustment operator, when (possibly) inconsistent probabilistic evidence is gathered.\footnote{With imprecise probabilities, inconsistency occurs when $\overline{P}(x)=0$ and positive evidence is provided  for some $x\in\Omega_X$.}

Let us start with simple probabilistic evidence: $P_X'$ on $\Omega_X$, such that $|\Omega_X|\geq2$. Imaginary kinematics can be introduced as a counterpart of PK for imaging.

\begin{definition}[Imaginary Kinematics]\label{gen:pk}
Any joint CS $K^\circ$ on $\mathbf{X}$ comes from $K$ by imaginary kinematics (IK) on a (possibly inconsistent) credal evidence $K_X'$ on variable $X$ whenever it holds:
\begin{description}
\item[IK1] $K^\circ(\alpha|x)\supseteq K_x^{\circ_I}(\alpha)$, for any $\alpha\in\Sigma$ and each $x\in\Omega_X$,
\item[IK2] $K^\circ(X) \models \Phi_X$, 
\item[IK3] $K^\circ(X)\equiv K_x^{\circ_I}(X)$ whenever $c_x=1$ for some $x\in\Omega_X$.
\end{description}
\end{definition}

Analogously, based on Definition~\ref{def:copk}, we provide an imaginary characterization of CPK defined as follows.

\begin{definition}[Imaginary Conditional Kinematics]
Let $K$, $K^\circ$ as above, such that $\underline{P}(y)>0$ for each $y\in\Omega_Y$. $K^\circ$ comes from $K$ on $\Omega_X\times\{Y=y\}$ based on imaginary conditional kinematics (ICK) if there exists a (possibly inconsistent) sequence $P_{X|y}'$ such that it holds:
\begin{description}
\item[ICK1] $K^\circ(\alpha|x,y)\supseteq K_x^{\circ_I}(\alpha|y)$, for each $x\in\Omega_X$,
\item[ICK2] $K^\circ(\alpha|y')\equiv K(\alpha|y')$, for each $y'\in\Omega_{Y}\backslash\{y\}$,
\item[ICK3] $K^\circ(Y)\equiv K(Y)$,
\item[ICK4] $K^\circ(x|y)\models \Phi_{X|y}$,
\item[ICK5] $K^\circ(X|y)\equiv K_x^{\circ_I}(X)$, whenever $c_x=1$, for some $x\in\Omega_X$.
\end{description}
\end{definition}

\section{Kinematical Imaginary Adjustment Rules}\label{ope}
For any $\alpha\in\Sigma$, if a CS $K$ over $\mathbf{X}$ is used to represent Your beliefs, imaging on $(\phi=\{x\})$ extends to:
\begin{equation*}
\left(K \circ_I \{x\}\right)(\alpha) = \left\{ P_x^{\circ_I}(\alpha) = \left(P \circ_I \{x\}\right)(\alpha), P\in K\right\}\,,
\end{equation*}
so that the lower envelope of $K$'s image on $\{x\}$, denoted as $K_x^{\circ_I}$, at $\alpha$, writes:
\begin{equation*}
\underline{P}_x^{\circ_I}(\alpha) = \min_{P(\mathbf{x})\in K(\mathbf{X})} \sum_{\mathbf{x}' \sim\alpha} \sum_{\mathbf{x}\in\Omega_\mathbf{X}}P(\mathbf{x}) \mathbb{I}_{\gamma(\mathbf{x},x)=\mathbf{x}'} \,.
\end{equation*}
By \cite[Th.1]{rens2016revision}, $K_x^{\circ_I}$ may be efficiently obtained by taking the convex hull (CH) of the images on $\{x\}$ of each $P \in \mathrm{ext}[K]$. Since the image of each $P\in \mathrm{ext}[K]$ at $\alpha=\{x'\}$ trivially corresponds to $P_x^{\circ_I}(x')=0$,\footnote{By definition, $P_x^{\circ_I}(x)=\sum_{\mathbf{x}\in\Omega}P(\mathbf{x})=1$.} whenever $x'\neq x$, refinement of $K_x^{\circ_I}(X)$ degenerates to a single PMF such that $P_X'(x)=1$, and zero otherwise. With a small abuse of notation, this yields the following:
\begin{equation*}
K_x^{\circ_I}(\mathbf{X})\equiv \left\{\begin{array}{ll}
1\cdot K(\mathbf{X}\backslash\{X\}) \quad \mathbf{x}\sim x\,,\\
0 \quad\quad\quad\quad\quad\quad \text{ otherwise.}\end{array}
\right.
\end{equation*}
\begin{example}\label{excred}
Let $K$ be a CS over $\mathbf{X}=\{X,Y\}$ specified by probability intervals as follows:
\begin{equation*}
K\begin{pmatrix}
x_1,y_1\\
x_1,y_2\\
x_2,y_1\\
x_2,y_2\\
x_3,y_1\\
x_3,y_2
\end{pmatrix}=
\begin{bmatrix}
0\\
0\\
0.15 - 0.35\\
0.25 - 0.49\\
0 - 0.45\\
0.03 - 0.5
\end{bmatrix}\,.
\end{equation*}
It is easy to see $\underline{P}_{x_1}^{\circ_I}(y_j)=\underline{P}(y_j)$, $j=1,2$, while $\overline{P}_{x_1}^{\circ_I}(x_k)=0$, $k=2,3$.
\end{example}

\subsection{Standard Probabilistic Evidence}
We start from the case of sharp probabilistic evidence on $\Omega_X$, i.e., $K_X'(X)=\{P_X'(X)\}$. The following adjustment operator extends Definition~\ref{def:ji}. As we did before for imaging, notation that is used with sharp beliefs applies to the generalized case of belief sets, when $|\mathrm{ext}[K]|\geq 1$.
\begin{definition}[(Probabilistic) Jeffrey's Imaging]\label{pji}
Let $K$ be any joint CS over $\mathbf{X}$ as above. Suppose probabilistic evidence $P_X'$ is provided over a (possibly) inconsistent collection of events, i.e., $\overline{P}(x)=0$, whereas $P_X'(x)>0$, for some $x\in\Omega_X$, $X\in\mathbf{X}$. For any event $\alpha$, $K_X^{\circ_{jI}}$ is the probabilistic Jeffrey's image of $K$ if it holds:
\begin{align*}
K_X^{\circ_{jI}}(\alpha)=\{&P_X^{\circ_{jI}}(\alpha)=\sum_{x\in\Omega_X}P_x^{\circ_I}(\alpha) P_X'(x),\\
&P_x^{\circ_I}\in K_x^{\circ_I}, x\in\Omega_X\} \,.
\end{align*}
That is, $K_X^{\circ_{jI}}(\alpha)=\left(K \circ_{jI} P_X'\right)(\alpha)$, for any $\alpha\in\Sigma$.
\end{definition}

The following result holds (the proofs of all the theorems are in the appendix).
\begin{theorem}\label{th:jipk}
Jeffrey's imaging is based on IK, and IK1 is strongly satisfied, i.e., $\models$ may be replaced by $\equiv$.
\end{theorem}

\begin{corollary}\label{cor}
Given sharp probabilistic knowledge on $\Omega_X$, the Jeffrey's image of any CS may be equivalently specified by the convexification of all PMFs $P^\circ$, each defined as follows:
\begin{equation*}
P^\circ(\alpha)=\sum_{x\sim\alpha}P_X'(x)P_i(\alpha)\quad \forall P\in \mathrm{ext}[K]\,.
\end{equation*}
\end{corollary}
It is easy to see that standard imaging is also trivially based on IK.

\begin{example}\label{ex1}
Consider the same setup as in Example~\ref{excred}, and suppose $P_X'(X)=\{(x_1,0.3), (x_2, 0), (x_3, 0.7)\}$. By Jeffrey's imaging on $P_X'$, we obtain $K_X^{\circ_{jI}}(Y)\equiv K(Y)$, while $\underline{P}_X^{\circ_{jI}}(y_j|x_i)\equiv K_{x_i}^{\circ_I}(y_j)$, $i=1,2,3$, $j=1,2$. Also, $K_X^{\circ_{jI}}(X)\models P_X'(X)$, and $K_X^{\circ_{jI}}$ is equivalent to the convex hull of PMFs $P^\circ$, defined as:
\begin{equation*}
P^\circ(x,y)=P_X'(x)P(y)\,,
\end{equation*}
for each $x\in\Omega_X,y\in\Omega_Y$ and $P\in \mathrm{ext}[K]$.
\end{example}

\subsection{Sharp Conditional Evidence}
We now introduce Adams' imaging as an adjustment operator $\circ_{aI}$, that extends $\circ_{jI}$ to the conditional case, just like revision rule $\circ_A$ extends $\circ_J$.

\begin{definition}[Adams' Imaging]\label{def:adamsim}
Let $K$ be any joint CS on $(\Omega, \Sigma)$ such that $\underline{P}(y)>0$, $Y\in\mathbf{X}$, and let conditional probabilistic knowledge $P_{X|y}'$ on $\Omega_X\times\{Y=y\}$. $K_{X|y}^{\circ_{aI}}$, the Adams' image of $K$ on $P_{X|y}'$, comes from $K$ by Adams' imaging $\circ_{aI}$, if it holds:
\begin{align*}
&K_{X|y}^{\circ_{aI}}(\alpha)=\\
&\{P_{X|y}^{\circ_{aI}}(\alpha)=P(\alpha,\neg y)+\sum_{x\in\Omega_X}P_x^{\circ_I}(\alpha,y)P_{X|y}'(x|y),\\
& P\in K, P_x^{\circ_I}\in K_x^{\circ_I}, x\in\Omega_X\}\,.
\end{align*}
I.e., $K_{X|y}^{\circ_{aI}}(\alpha)=\left(K \circ_{aI} P_{X|y}'\right)(\alpha)$, for any $\alpha\in\Sigma$.
\end{definition}
When $|\mathrm{ext}[K]|=1$, from previous considerations, Adams' imaging reduces to the following:
\begin{equation}\label{adamsprec}
P_{X|y}^{\circ_{aI}}(\alpha)=P(\alpha,\neg y)+\sum_{x\in\Omega_X}P_x^{\circ_I}(\alpha,y)P_{X|y}'(x|y)\,.
\end{equation}
\begin{example}[Ex.~\ref{celeste} continued]
The Adams' image on $P_{X|y}'$ of Celeste's beliefs on $\Omega_X\times\Omega_Z$ is the following:
\begin{equation*}
P_{X|y}^{\circ_{aI}}
\begin{pmatrix}
x_Wz\\
x_W \neg z\\
x_Gz\\
x_G\neg z\\
x_Bz\\
x_B\neg z\\
\end{pmatrix}=
\begin{bmatrix}
0.5620\\
0.1480\\
0.0845\\
0.0755\\
0.0785\\
0.0515
\end{bmatrix} \,.
\end{equation*}

It holds $P_{X|y}^{\circ_{aI}}(X|y)=P_{X|y}'(X|y)$ and $P_{X|y}^{\circ_{aI}}(Y,Z)=P(Y,Z)$. Adjustment of her beliefs by $P_{X|y}'$ yields $P_{X|y}^{\circ_{aI}}(z|x_B)\approx0.6$, whereas $P(z|x_B)=0.2$. Thus, Celeste rapidly swims back to shore.
\end{example}
As a remark, inconsistency of $P_{X|y}'(x|y)$, for some $x\in\Omega_X$, with respect to any PMF $P$, may refer to either i) $P(x|y)=0$, while $P(y)>0$, (this is just the case of Adams' imaging above), or ii) $P(y)=0$ in the first place, and possibly $P(x|y)=0$. We argue case ii) deserves some caution, since full inconsistency of event $(Y=y)$ is likely not to yield any further conjecturing on related events, from a modeler's perspective. E.g., You are certain that no alien lives on Mars. Is it worth include Your belief on the alien having long hair in Your belief base, provided that You are not admitting the alien's existence upstream? On the other hand, we reckon arguments may be easily raised against our position, starting from our proposed running example. Still, if no evidence is provided on $\Omega_Y$, a cautious approach would require application of an iterated procedure. We leave this point for future work.

It is now straightforward to note that Adams' imaging generalizes Jeffrey's imaging to the conditional setting. 

\begin{theorem}\label{th:aim}
Adams' imaging is based on ICK, and ICK1 is strongly satisfied. Eq.~\eqref{adamsprec} strongly satisfies all conditions.
\end{theorem}

Analogously to Corollary~\ref{cor}, it might be easily shown that $K_{X|y}^{\circ_{aI}}$ at any $\mathbf{x}\sim y$ is equivalent to the CS obtained taking the product of sharp assessment $P_{X|y}'$ and the marginalization over variable $X$ of the original belief set $K$. We also provide the following additional result, which extends \cite{rens2016revision}.

\begin{theorem}
Both Jeffrey's and Adams' imaging satisfy consistency axioms KM1, KM3 and KM4. KM2, KM5 and KM6 are satisfied only is $K$ is degenerate at $(X|y)$, i.e., $|K(X|y)|=1$ (and at $(Z|w)$, for KM5 and KM6).
\end{theorem}

\subsection{Credal Jeffrey's Imaging}
When beliefs are expressed as a joint CS over $\mathbf{X}$, adjustment by a single reliable PMF requires simultaneous computation of all bounds spanned by the updating of each $P\in K$. Also in this case, adjustment may be restricted to the PMFs in $\mathrm{ext}[K]$ only, and their convex hull consequently considered.
\begin{definition}[Credal Jeffrey's Imaging]\label{def:cjr}
Given CS $K$ over $\mathbf{X}$ and credal probabilistic evidence $K_X'(X)$, we define credal Jeffrey's imaging $\circ_{cjI}$ as the functional mapping $K$ to CS $K_X^{\circ_{cjI}}$, consistent with $K_X'(X)$ as follows:
\begin{align*}
&K_X^{\circ_{cjI}}(\alpha)\\&=\left\{P^\circ(\alpha)=\left(P \circ_{jI} P_{X}'\right)(\alpha), \begin{array}{ll}
P(\mathbf{X})\in K(\mathbf{X}), \\
P_X' \in K_X'(X)
\end{array}
\right\}
\end{align*}
\end{definition}

The following result generalizes Theorem~\ref{th:jipk}.

\begin{theorem}\label{th:credalimaginaryj}
Given (possibly) inconsistent credal probabilistic evidence, credal Jeffrey's imaging yields the unique joint CS based on IK.
\end{theorem}

\begin{table}[htp!]
\caption{Summary of belief adjustment rules/properties.}
\label{table:rulesprecise}
\begin{center}
\begin{tabular}{ccc}
\hline
\textbf{RULE}	& $\Phi_*$ & \textbf{KINEMATICS}\\
\hline 
$\circ_J$	& $\phi_x=c_x, \forall x$	& PK\\
$\circ_A$	& $\{y\to x\}=c_x, \forall x$	& CPK\\
$\circ_{jI}$	& $\{x\}=c_x, \forall x$	& IK (Th.~\ref{th:jipk}) \\
$\circ_{aI}$	& $\{y\to x\}=c_x, \forall x$	& ICK (Th.~\ref{th:aim}) \\
$\circ_{cjI}$	& $\{x\}\bowtie c_x, \forall x$	& IK (Th.~\ref{th:credalimaginaryj})\\
\hline
\end{tabular}
\end{center}
\end{table}

\section{Conclusions and Future Work}
We introduced adjustment operators based on Lewis' imaging functional, to deal with probabilistic inconsistent evidence, in a generalized setting of imprecise probabilities, specified by credal sets. These are summarized in Table~\ref{table:rulesprecise}. We point out that the revision rules (conditioning, Jeffrey's rule and Adams' conditioning) are not fully general due to partiality, whereas the remaining succeed in adjusting a given belief set following inconsistent observations. 

Further generalization to the case of credal conditional probabilistic evidence is not straightforward as the adjustment process would likely incur in dilating mechanics, resulting in detrimental loose inclusion relationships. This reasoning also applies to the iterated framework, where additional considerations must be formulated on the role evidence plays on the adjustment process. As a future work we will tackle this sort of scenarios. Besides that, we also intend to compare our approach against methods based on lexicographic probabilities (e.g., \cite{benavoli}) as well as applying these ideas to probabilistic graphical models by extending what have been already done for Jeffrey's rule in \cite{antonucci}.
\appendix
\section{Proofs}
This appendix provides proofs to the results stated in the paper.
\begin{proofoft}
To prove $\circ_{jI}$ is based on IK, we must check it produces a CS that satisfies IK1-IK3. Motivated by \cite[Th.1]{rens2016revision}, we restrict our attention toward the extreme points of $K$. Without loss of generality, let $\mathbf{X}=\{X,Y\}$. Each extreme point of $K(\mathbf{X})$, say $P_{j,k}\in \mathrm{ext}[K]$, may be equivalently specified as:
\begin{equation}\label{fact}
P_{j,k}(x,y)=P(x|x'_j)P(y|x,y_k')\,,
\end{equation}
with $P(y|x,y_k')$ is set equal to zero whenever it is undefined and $P(x|x_k')=0$.\footnote{As a remark, $P(x)=0$ does not necessarily imply $P(y|x)=0$, in De Finetti's view.} $X'$ and $Y'$ are uniformly distributed auxiliary random variables, used to index $K$'s extreme points at $X$ and at $Y|X$, respectively. This way, for a given ordering,
\begin{align*}
P(x|x'_1)&= \sum_{y_k',y}P(x|x_1')P(y|x,y_k')P(y_k')\\
&=\underline{P}(x)\,,
\end{align*}
and $\underline{P}(x,y)=P(x|x_1')P(y|x,y_1')$.\\
It holds:
\begin{align*}
\underline{P}_x^{\circ_I}(x)&=P_x^{\circ_I}(x|x_1')\\
&= \sum_{y_k',y,x} P(x|x_1')P(y|x,y_k')P(y_k')\\
&\leq1\,.
\end{align*}
Since $P_x^{\circ_I}(x'|x_1')=0$, for any $x'\neq x$ in $\Omega_X$, refinement of $K_x^{\circ_I}(x)$ degenerates at 1. If $P_X'\models (\phi=\{x\})$, IK3 is satisfied.\\
When a non-trivial PMF is provided, i.e., $P(x)>0$ for at least two elements in $\Omega_X$, it holds:
\begin{align*}
P_X^{\circ_{jI}}(x|x_1')&=\left[\sum_{y_k',y,x}P(x|x_1')P(y|x,y_k')P(y_k')\right]P_X'(x)\\
&\leq P_X'(x) \,,
\end{align*}
and similarly $P_X^{\circ_{jI}}(x|x_{|\mathrm{ext}[K]|}')\geq P_X'(x)$. This proves IK2 since $K_X^{\circ_{jI}}(X)\ni P_X'(X)$.\\
Proof of IK1 is also straightforward:
\begin{align*}
P_X^{\circ_{jI}}(y|x,y_1')&= \frac{\left[\sum_{x,x_j'}P(x|x_j')P(y|x,y_1')\right]P_X'(x)}{P_X'(x)}\\
&= \sum_{x,x_j'}P(x|x_j')P(y|x,y_1')\\
&= P_x^{\circ_I}(y|y_1')
\end{align*}
Analogous reasoning applies to the upper envelope, and thus $K_X^{\circ_{jI}}(Y|x)\equiv K_x^{\circ_I}(Y)$. This ends the proof.
\begin{flushright}$\square$\end{flushright}
\end{proofoft}

\begin{proofoft}
To prove $\circ_{aI}$ is based on ICK we need to check ICK1-ICK5 are satisfied by $K^\circ=\left(K\circ_{aI} P_{X|y}'\right)$. When $|\mathrm{ext}[K]|=1$, ICK1-ICK5 reduce to the following:
\begin{description}
\item[ICK1'] $P^\circ(\alpha|x,y)=P_x^{\circ_I}(\alpha|y)$, for each $x\in\Omega_X$,
\item[ICK2'] $P^\circ(\alpha|y')=P(\alpha|y')$,
\item[ICK3'] $P^\circ(Y)=P(Y)$,
\item[ICK4'] $P^\circ(X|y)=P_{X|y}'(X|y)$,
\item[ICK5'] $P^\circ(X|y)=P_x^{\circ_I}(X|y)$, whenever $P_{X|y}'(x|y)=1$ for some $x\in\Omega_X$.
\end{description}
We first prove consistency points ICK4' and ICK5'. Let $P_{X|y}'$ be any PMF on $\Omega_X\times\{Y=y\}$, it holds:
\begin{align*}
P_{X|y}^{\circ_{aI}}(x|y)&= \frac{P_x^{\circ_I}(y)P_{X|y}'(x|y)}{\sum_xP_x^{\circ_I}(y)P_{X|y}'(x|y)}\\
&= P_{X|y}'(x|y)
\end{align*}
since $P_x^{\circ_I}(x,y)=P_x^{\circ_I}(y)=P(y)$, whatever $x\in\Omega_X$. Also, $\sum_xP_{X|y}'(x|y)=1$ by definition. If $P_{X|y}'(x|y)=1$ for some $x$, $P_{X|y}^{\circ_{aI}}(x|y)=1$, 0 otherwise. The following holds:
\begin{align*}
\underline{P}^\circ(x|y)&=\min_{P^\circ \in \mathrm{ext}[K^\circ]}P^\circ(x|y)\\
&=P_{X|y}'(x|y)\frac{\overline{P}_x^{\circ_I}(y)}{\underline{P}_x^{\circ_I}(y)}\\
&\leq P_{X|y}'(x|y)\,.
\end{align*}
Similarly, $\overline{P}^\circ(X|y)\geq P_{X|y}'(X|y)$, for each $P^\circ\in \mathrm{ext}[K^\circ]$.\\
We now prove condition ICK1 (and thus ICK1') is satisfied by $\circ_{aI}$. Without loss of generality, let $\mathbf{X}=\{X,Y,Z\}$. It holds:
\begin{align*}
\underline{P}^\circ(z|x,y)&= \frac{P_{X|y}'(x|y)\underline{P}_x^{\circ_I}(z,y)}{P_{X|y}'(x|y)\overline{P}_x^{\circ_I}(y)}\\
&=\underline{P}_x^{\circ_I}(z|y)\,.
\end{align*}
As for point ICK2 (and ICK2'), it trivially holds by Definition~\ref{def:adamsim}:
\begin{align*}
\underline{P}^\circ(z|y')&=\underline{P}(z|y') \,.
\end{align*}
for any $y'\neq y$. ICK3' is proved analogously, since $P_{X|y}^{\circ_{aI}}(y)=1-P(\neg y)=1-\sum_{y'\neq y}P_{X|y}^{\circ_{aI}}(y')$. Similarly, fulfillment of ICK3 may be derived by the conjugacy relation \cite{walley1991statistical}.
\begin{flushright}$\square$\end{flushright}
\end{proofoft}
\begin{proofoft}
Consider CS $K$ and conditional probabilistic evidence $P_{X|y}'(X|y)$. To avoid cumbersome notation, we write $\circ$ to denote $\circ_{aI}$ throughout the proof. Also, we refer to general formula $\phi=c$ to denote both $\phi_x$ and $\phi_{x|y}$.\\
KM1 and KM3 follow from IK2 and ICK4 (cfr Th.1 and Th.2, respectively).\\
We prove KM2 is not satisfied under general conditions. Consider the lower envelope of $K$ at $(x|y)$. If $K\models P_{X|y}'$, it holds:
\begin{align*}
\underline{P}(x|y)&\leq P_{X|y}'(x|y)
\end{align*}
by definition, and $\left(K\cup P_{X|y}'\right)=K$. From previous discussion, we expect $(K \circ P_{X|y}') \supseteq K$, equality holding if and only if $K(\mathbf{X})$ may be equivalently specified as the product of sharp conditional assessment on $\Omega_X\times\{Y=y\}$ and CS over $\left(\mathbf{X}\backslash\{Y\},y\right)$. Same reasoning applies to KM5 and KM6. These three postulates are satisfied if and only if $K$ is already degenerate at the domain of probabilistic evidence, and consistent with it already.\\
Postulate KM4 holds by \cite[Th.1]{rens2016revision}.
\begin{flushright}$\square$\end{flushright}
\end{proofoft}
The following preliminary result holds:
\begin{lemma}\label{inclusion}
Let $K$ be a joint CS over $\mathbf{X}$, and let $K_X'$ denote a credal probabilistic finding, gathered on $\Omega_X$. For any event $\alpha$, the Jeffrey's image $K_X^{\circ_{cjI}}(\alpha)$ of $K(\alpha)$ on $K_X'(X)$ satisfies the following:
\begin{equation*}
K_X^{\circ_{cjI}}(\alpha)\supseteq K_X^{\circ_{cjI}}(\alpha|x)\supseteq K_x^{\circ_I}(\alpha)
\end{equation*}
for any $\alpha\in\Sigma$. Equality holds when $|K'(X)|=1$.
\end{lemma}
\begin{proofofl}
Let $\mathbf{X}=\{X,Y\}$ and $K$ be any CS over $\Omega$. $K_X'$ is gathered on $\Omega_X$, to adjust $K$ accordingly. By definition of credal Jeffrey's imaging, it holds:
\begin{align*}
\min_{P_X^{\circ_{cjI}}\in K_X^{\circ_{jI}}}P_X^{\circ_{cjI}}(y|x)&=\min_{P(y)\in K_(Y)}P(y)\frac{\underline{P}_X'(x)}{\overline{P}_X'(x)}\\
&\leq \min_{P(y)\in K(Y)}P(y) \,,
\end{align*}
and analogously for the upper envelope, with $\geq$. This proves the rightest inclusion relationship: $K_X^{\circ_{cjI}}(Y|x)\supseteq K_x^{\circ_I}(Y)(\equiv K(Y))$.\\
We now prove inclusion of $K_X^{\circ_{cjI}}(y|x)$ by $K_X^{\circ_{cjI}}(y)$:
\begin{align*}
\frac{\underline{P}_X^{\circ_{cjI}}(y)}{\underline{P}_X^{\circ_{cjI}}(y|x)}&=\frac{\underline{P}(y)\sum_x\underline{P}_X'(x)}{\underline{P}(y)\frac{\underline{P}_X'(x)}{\overline{P}_X'(x)}}\\
&=\overline{P}_X'(x)\sum_{x'\neq x}\underline{P}_{X}'(x')\\
&\leq 1 \,.
\end{align*}
Hence $\underline{P}_X^{\circ_{cjI}}(y)\leq \underline{P}_X^{\circ_{cjI}}(y|x)$, for any $x\in\Omega_X$, $y\in\Omega_Y$. $\overline{P}_X^{\circ_{cjI}}(y)\geq \overline{P}_X^{\circ_{cjI}}(y|x)$ is derived analogously.\\
Equality holds when $K_X'(X)=\{P_X'(X)\}$ as $\underline{P}_X'(x)=\overline{P}_X'(x)$, for each $x\in\Omega_X$, summing to one.
\begin{flushright}$\square$\end{flushright}
\end{proofofl}
\begin{proofoft}
Given a joint CS $K$ over $\mathbf{X}$ and $K_X'$, let $\circ$ denote credal Jeffrey's imaging.\\
IK1 is satisfied by Lemma~\ref{inclusion}. IK2 is also satisfied as it holds:
\begin{align*}
\underline{P}_X^{\circ_{cjI}}(x)&= 1 \cdot \underline{P}_X'(x)\,,
\end{align*}
for each $x\in\Omega_X$. And analogously for $\overline{P}_X^{\circ_{cjI}}(X)$. When $K_X'(X)=\{P_X'(X)\}$ such that $P_X'(x)=1$, IK3 is satisfied since $\circ_{cjI}$ reduces to $\circ_{jI}$.
\begin{flushright}$\square$\end{flushright}
\end{proofoft}
\bibliographystyle{amsplain} 
\bibliography{bibfile.bib}
\end{document}